\documentclass[letterpaper, 10 pt, conference]{ieeeconf}

\IEEEoverridecommandlockouts
\usepackage{graphics} 
\usepackage{amsmath} 
\usepackage{amssymb}  
\usepackage{gensymb}
\usepackage[table]{xcolor}
\usepackage[ruled,linesnumbered, noend]{algorithm2e}
\usepackage{url}

\usepackage{soul}
\usepackage{cite}
\usepackage{comment}
\usepackage{multirow}
\usepackage{graphicx}
\usepackage{wrapfig}
\usepackage{subfigure}
\usepackage{balance}
\usepackage{soul, color}
\usepackage{wrapfig}
\usepackage{diagbox}
\usepackage{adjustbox}

\usepackage{amsthm}

\newcommand{\ve}[1]{\mathbf{#1}} 
\newcommand{\tve}[1]{\tilde{\mathbf{#1}}} 

\newcommand{\n}[0]{PNG }
\newcommand{\nm}[0]{PNG}
\newcommand{\nn}[1]{PNG-{#1}}

\title{Print-N-Grip: A Disposable, Compliant, Scalable and One-Shot 3D-Printed Multi-Fingered Robotic Hand}

\author{Alon Laron$^1$, Eran Sne$^2$, Yaron Perets$^2$ and Avishai Sintov$^1$
\thanks{A. Laron and A. Sintov are with the School of Mechanical Engineering, Tel-Aviv University, Israel. e-mail: {\small alonlaron@mail.tau.ac.il, sintov1@tauex.tau.ac.il}.}
\thanks{E. Sne and Y. Perets are with the Nuclear Research Center - Negev, Israel.}
}

\begin{document}


\maketitle
\thispagestyle{empty}
\pagestyle{empty}

\begin{abstract}
{\it Robotic hands are an important tool for replacing humans in handling toxic or radioactive materials. However, these are usually highly expensive, and in many cases, once they are contaminated, they cannot be re-used. Some solutions cope with this challenge by 3D printing parts of a tendon-based hand. However, fabrication requires additional assembly steps. Therefore, a novice user may have difficulties fabricating a hand upon contamination of the previous one. We propose the \textit{Print-N-Grip} (PNG) hand which is a tendon-based underactuated mechanism able to adapt to the shape of objects. The hand is fabricated through one-shot 3D printing with no additional engineering effort, and can accommodate a number of fingers as desired by the practitioner. Due to its low cost, the PNG hand can easily be detached from a universal base for disposing upon contamination, and replaced by a newly printed one. In addition, the PNG hand is scalable such that one can effortlessly resize the computerized model and print. We present the design of the PNG hand along with experiments to show the capabilities and high durability of the hand.}

\end{abstract}


\section{Introduction}
\label{sec:introduction}

The usage of robots in factories is commonly limited to dry tasks in order to preserve the integrity of expensive hardware. Contact of robotic hands with liquids may cause corrosion in critical moving parts and short circuit in electrical components \cite{Siciliano2007}. Corrosion from other chemicals, such as acids in chemical plants \cite{Huang2021} or pesticides in agriculture \cite{Lojans2012}, may be even worse yielding rapid decay. In order to prevent damage of the hand in wet conditions, waterproofing is essential \cite{Meng2006}. As such, the choice of non-corrosive materials and sealing components add extra costs to the robotic hardware. Other risks include dusty tasks where small particles can abrade and clog moving parts. In radioactive environments, electronic components may become immediately inoperable upon exposure to radiation \cite{Garg2006}. The common approaches to cope with radiation are to either use standard and replaceable electric components or to avoid them entirely \cite{Zhang2020}. In conclusion, extreme conditions may yield extensive wear and tear of the robotic hardware, significantly decrease its life-time and lead to financial loss. In addition, some chemical (e.g., pesticides, acids and radioactive materials) and biohazard (e.g., viruses and microbes from medical procedures) materials often cannot be cleaned such that the expensive hardware would have to be disposed \cite{Siciliano2007}. 
\begin{figure}
    \centering
    \includegraphics[width=\linewidth]{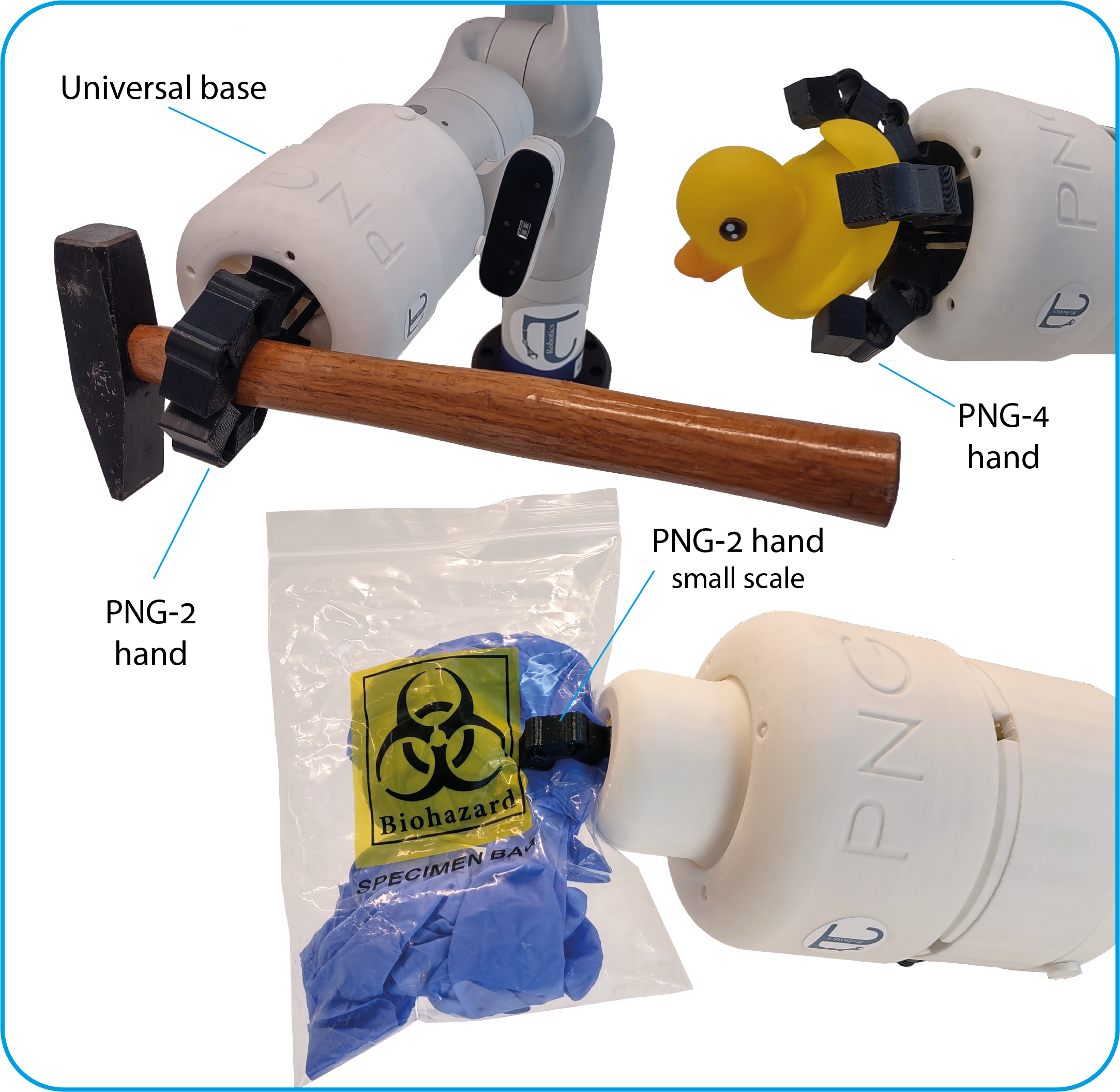}
    \caption{Three variations of the Print-N-Grip (\nm) hand: (top left) two-finger, (top right) four-finger and (bottom) small scale two-finger grippers mounted on a universal base and grasping a hammer, rubber duck and a bio-hazard bag, respectively.}
    \label{fig:intro}
    \vspace{-0.6cm}
\end{figure}
\begin{figure*}
    \centering
    \includegraphics[width=\linewidth]{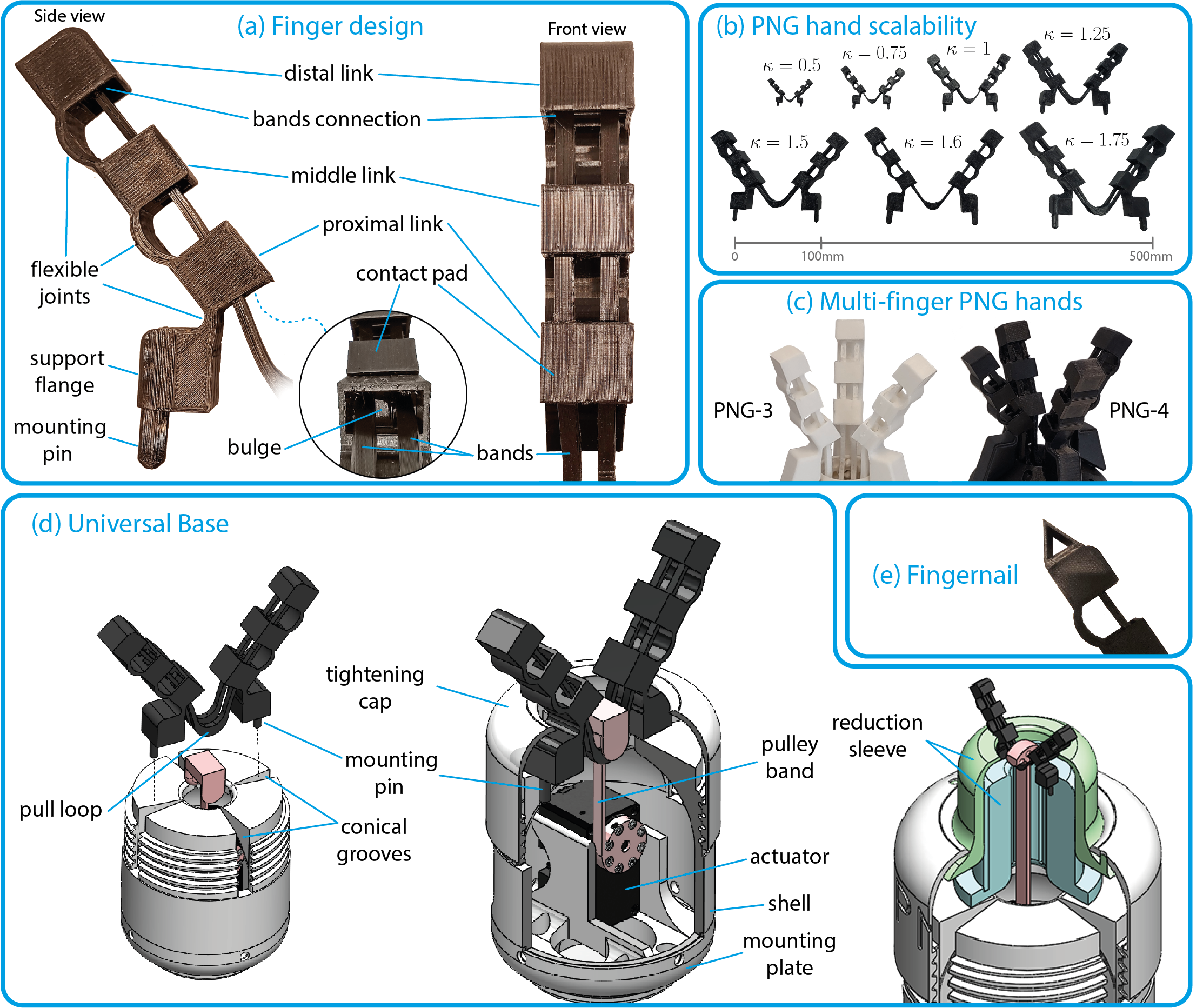}
    \caption{(a) Front and side view of a \n finger. Internal view of the proximal link is seen in the circle where the two bands are separated by a bulge. Due to its height, the bulge prevents from the contact pad to press onto the bands. (b) Different scales $\kappa$ of a \nn{2} hand. As convention, a hand with a finger length of 52~mm is defined as $\kappa=1$. Correspondingly, the smallest and largest hands shown and evaluated are with finger lengths of 25.6~mm ($\kappa=0.5$) and 89.7~mm  ($\kappa=1.75$), respectively. (c) Examples of three- (\nn{3}) and four- (\nn{4}) finger hands. Both hands are one-shot 3D printed. (d) Universal base for mounting the printed hand. The 3D printed hand is fixed to the base by inserting the mounting pins into the conical grooves and closing with the tightening cap. A reduction sleeve is used for hands of small scale. (e) Fingernail on the distal link assists in picking-up small objects.}
    \label{fig:main}
    \vspace{-0.5cm}
\end{figure*}

Compliant hands are underactuated mechanisms that utilize tendons and exhibit the ability to passively adapt to objects of uncertain size and shape \cite{Salvietti2017}. Hence, they have gained popularity in the past two decades by enabling a low-cost and compact design compared to conventional robotic hands. By definition, underactuated hands have more joints than actuators. Each finger has compliant joints by either including springs or flexible bands. An actuator provides flexion through a tendon running along the length of each finger. Consequently, joints in a finger are coupled together yielding an adaptability property to an object's shape. Therefore, a robotic hand with such fingers can provide a stable and robust grasp without tactile sensing or prior planning, and with open-loop control \cite{Odhner2015}.  Underactuated, or compliant, hands can provide robust and efficient performance for real-world applications in, for instance, service and household robots \cite{Kim2014}, warehouse sorting and bin picking \cite{Kragten2012} and flexible industrial automation \cite{Negrello2020}. 

Advances in 3D printing technology have simplified the fabrication of such hands making them more accessible and low-cost \cite{Ma2017YaleOP}. Computer-Aided Design (CAD) files can be freely downloaded, 3D printed and assembled along with other unprintable hardware such as actuators, pins and springs. Joints require either the installation of springs or molding of urethane rubber. Therefore, non-expert or novice users will still have difficult time assembling the hand as it requires technical skills and extensive work time. In addition, threading the tendons along the fingers is a tedious task and they tend to tear frequently. Since the tendons can tear, tension forces on the tendons must be bounded yielding limited ability to counter-act external forces on the grasped object. Hence and due to wear, the hands require frequent time-consuming maintenance.

In this work, we propose a novel robotic hand, termed the \textit{Print-N-Grip} (\nm) hand, that is designed for a wide range of applications, but is particularly suited for use in wet industries. Excluding a universal base, the \n hand is fabricated entirely by one-shot 3D printing. Hence, the fingers, joints and tendons are all printed as one unit with a flexible Thermoplastic Polyurethane (TPU) filament. The unique design includes bands that operate as tendons printed with the hand and remove the need for tedious threading of tendon strings. Therefore, the hand is low-cost and durable. The \n hand, seen in Figure \ref{fig:intro}, can accommodate several fingers and we explore and demonstrate the use of two-, three- and four-finger hands. In addition, it is shown that any \n hand is scalable. In other words, the hand can be printed in a different scale corresponding to the intended task. For example, the hand can be miniaturized for medical applications or printed bigger for large material handling, with minimal effort. 

Since the \n hand is fully printed with no electronics, the hand can be considered disposable. A  worker, for example, in a chemical or biological factory can easily replace it once contaminated. The actuator at the universal base does not come in contact with material. Hence, upon contamination of the hand, only the hand is replaced and the base remains clean. A newly \n hand is immediately ready for use after 3D printing and no technical assembly is required expect from simple mounting on the universal base. In order to cope with environmental issues of disposing and since no electronics are on the hand, a disinfection or decontamination process can be devised in order to reuse the hand. However, such process is not in the scope of this work.

For printing the hand, no expensive or professional 3D printer is required and a simple Fused Deposition Modelling (FDM) one is sufficient. In this work, we present the design and structure of the \n hand along with a mechanical model of a finger. Then, the model is experimentally validated. Following, additional experiments show the lifting capacity of \n hands of different scale. Furthermore, we explore the ability of various \n hands with various number of fingers and scales to pick-up different everyday objects. Detailed structure and additional properties of the \n hand are seen in Figure \ref{fig:main}. In conclusion, the \n hand provides a low-cost and compliant solution for robotic operation in harsh and contaminated environments with minimal maintenance effort.

\section{Related work}
\label{sec:related_work}

One can make a distinction between two common types of tendon-based robotic hands: anthropomorphic and non-anthropomorphic hands. Anthropomorphic robotic hands are highly articulated humanoid hands which have been the topic of research for a considerable period of time \cite{Salisbury1982}. They commonly have three or four fingers opposing a thumb. Some anthropomorphic hands, such as the Shadow \cite{Tuffield2003} and DLR \cite{Grebenstein2012}, are based on tendons. While they are dexterous and have been demonstrated in complex tasks, they are highly expensive, demand significant effort to calibrate and maintain, require active sensing and their control can be complex even in simple grasping tasks. Consequently, few are actually used in real-world applications. 

Non-anthropomorphic tendon-based hands are multi-finger designs that do not attempt to mimic the human hand. In other words, non-anthropomorphic hands are thumb-less with unnatural design \cite{Ramos1999}. Parallel jaw grippers, for instance, are widely used due to their simplicity and
durability. They are ubiquitous in industrial application of material handling
\cite{Guo2017}. However, jaw grippers normally have only one DOF and are rigid such that they cannot adapt to the shape of an object. Hence, they require precise pick-up and closing control for not damaging the object. A prominent class of non-anthropomorphic hands is, therefore, compliant hands. Early compliant hands like the Shape Deposition Manufacturing (SDM) \cite{Dollar2010} and iRobot-Harvard-Yale (iHY) \cite{odhner2014compliant} hands have demonstrated the ability of such designs to perform both power and pinch grasping. Both have led to the commercial design of the three-finger ReFlex hand by Righthand Robotics \cite{Littlefield2016}.

One-shot 3D printing is not a novel concept. In particular, monolithic mechanisms have existed for quite some time \cite{Smith1987, Kota2001}. However, usages are mostly focused on high-precision positioning \cite{Culpepper2004,Choi2006,Qin2014,Dao2017}. Yet, not much prior work exists on one-shot 3D printing of robotic hands. Of the few, recent work have proposed a one-shot printed granular jamming gripper \cite{Howard2022}. Another work focused on one-shot fabrication of a finger while embedding tactile sensors and shape memory alloy actuator \cite{Stano2022}. A compliant parallel gripper was proposed and designated for extracting solid objects from powder \cite{Cormack2021}. Similarly, a bio-inspired parallel gripper was proposed where the joint of each finger receives its compliance from a spiral spring \cite{Zolfagharian2022}. In \cite{Velasquez2023}, a computational approach was presented for designing a one-shot 3D printed two-finger gripper with bi-inspired and sliding joints. RetracTip is a sea-anemone-inspired gripper formed by a bi-stable membrane to achieve delicate grasping \cite{Qi2022}. The mechanism is fabricated in a multi-material and high-cost 3D printer (e.g., Stratasys). While innovative, the proposed grippers are not tendon-based and cannot offer high capacity lifting and durability.

A more recent work introduced a one-shot 3D printed two-finger gripper which operates on pneumatic pressure \cite{Zhai2023}. A work most related to ours proposed a one-shot printed hand with an integrated tendon line \cite{Cormack2020}. Fabrication of the hand requires a high-cost multi jet fusion printer and, due to the thin tendon, cannot provide high-capacity power grasping or scalability. In addition, it does not enable rapid and easy replacement using a universal base. While not one-shot, a recent and intriguing work has proposed the two-finger underactuated InstaGrasp gripper \cite{Zhou2023}. The gripper has the structure of the OpenHand \cite{Ma2017YaleOP} Model T42 where each finger is composed of nine 3D printed parts of Polylactic Acid (PLA) and TPU. While almost entirely 3D printed and compliant, the hand requires some assembly. Unlike the proposed PNG hand in this paper, these 3D printed robotic hands do not demonstrate high-load grasping, are not scalable and do not offer easy replacement for disposing if contaminated.

A different type of compliant hands worth mentioning is the soft hands \cite{Deimel2013,Homberg2015,Teeple2021}. They are usually made of soft and pliable materials, such as silicone or elastomers, which allow them to passively conform to the shape of the object. Hence, the objects can be gripped securely without damage. Soft robotic hands are often used in applications where a gentle touch is required. They are commonly operated through pneumatic actuation. RBO Hand 2 is a  five-finger soft and compliant hand where each finger acts as a pneumatic continuum actuator \cite{Deimel2016}. Inelastic strings warp each finger such that, when inflated, pressure forces elongate them in desired directions. While promising and potentially low-cost, they are not strong or durable as hands with rigid structure which limits their ability to lift heavy objects. Also, they can be more difficult to control due to sensitivity to changes in pressure and temperature.







\section{Design}
\label{sec:Design}

In this section, we present the full design of the PNG hand starting with the structure of a finger, followed by the design of various multi-finger hands and their universal base.

\begin{figure}
    \centering
    \begin{tabular}{cc}
         \includegraphics[width=0.49\linewidth]{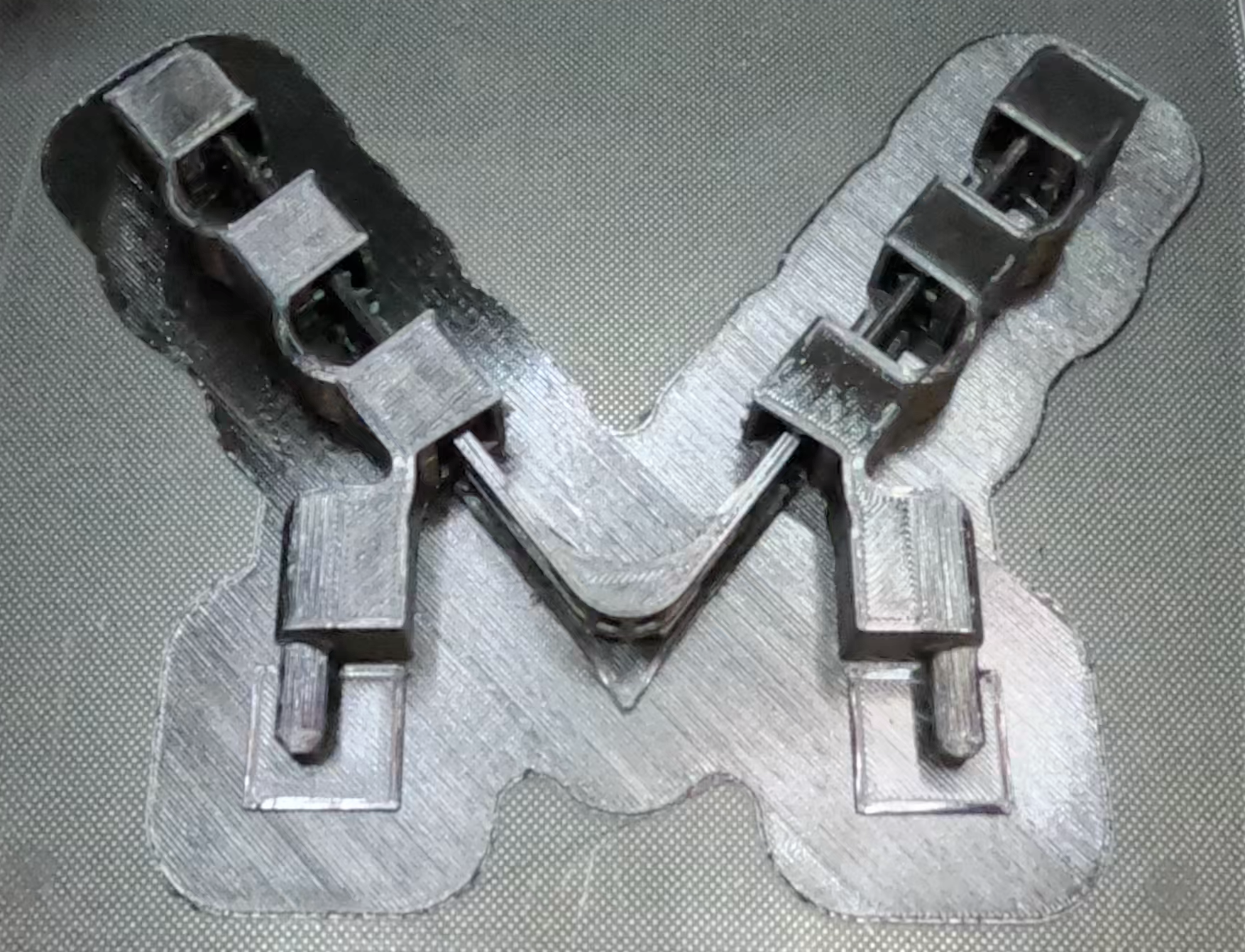} \hspace{-0.5cm} &  \includegraphics[width=0.49\linewidth]{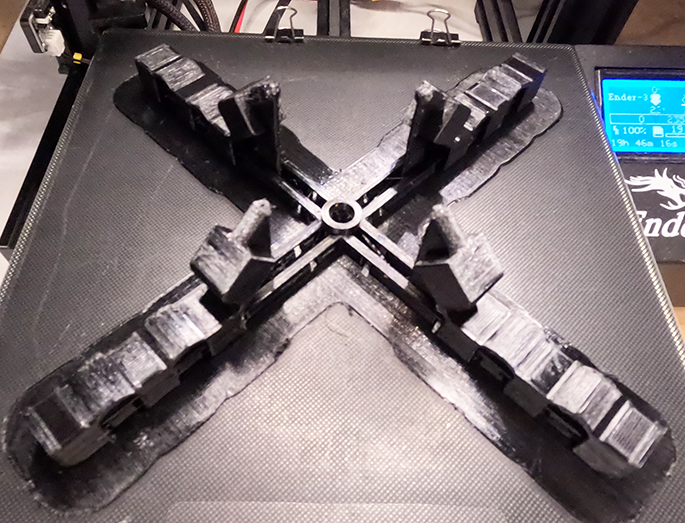} \\
         (a) & (b) \\
    \end{tabular}
    \caption{(a) \nn{2} and (b) \nn{4} hands on the 3D printer bed after printing is completed.} 
    \label{fig:bed}
\end{figure}

\subsection{Finger design}
\label{sec:finger_design}

The design of the finger is inspired by common formations such as in the SDM hand \cite{Dollar2010}, the OpenHand project \cite{Ma2017YaleOP} and an anthropomorphic hand \cite{Gao2021}. However, these hands require assembly of several parts and the installation of tendon wires. The proposed finger design is printed as one unit and does not require any assembly. A PNG finger is composed of four links: support flange, proximal link, middle link and distal link, as seen in Figure \ref{fig:main}a. Each of the latter three has a contact pad which interacts with the object. The support flange, with assistance of the mounting pin, is mounted on the base unit and will be discussed later. Each two links are connected by a flexible joint. While the finger can be designed with more than three joints, prior work supports this configuration and preliminary analysis has shown that this form is sufficient for power grasping of various objects \cite{Dollar2010}. Nevertheless, practitioners can modify the finger to include a different number of links and joints. In addition, a fingernail, as seen in Figure \ref{fig:main}e, can be included on the distal link in order for the hand to pick-up small objects such as a coin or flash drive.

Since the finger is flexible, a single tendon running along the center of finger may cause it to twist when pulled. In addition, contact of a link with an object would apply pressure on the tendon and lead to large frictions. Therefore, replacing the common threaded tendon, two band tendons run along the finger within the proximal and middle links, and connected to the distal link. The two tendons are separated by a bulge higher than the thickness of the bands as seen at the bottom of Figure \ref{fig:main}a. In this formation, the bulge prevents the contact pad from exerting undesired forces on the bands. The bands become one wide band towards the support flange such that they are pulled in tandem for stable flexion. After 3D printing, the bands are connected to the links by thin support bridges which are imperative for successful fabrication. When the tendons are pulled while the support flange is fixed, they slide within the links and the finger flexes. Hence, the support bridges are torn in the first flexion cycle. Upon release of tendon force, the elasticity of the flexible joints enables extension of the finger to its resting position. 

\subsection{Multi-finger hand}
\label{sec:multi_hand}

A PNG hand is printed with $n$ fingers in one-shot with a standard FDM printer. For simple notation, we denote an $n$-finger PNG hand as \nn{$n$}, e.g., \nn{3} denotes a three-finger PNG hand. Thermoplastic Polyurethane (TPU) filament is a flexible and strong material and, thus, used for the 3D printing. TPU has good corrosion resistance such that chemicals have minor effect \cite{Liu2019}. Note that TPU has a heat resistance of up to approximately 80$^\circ$C. Hence, the hand can go through multi-use sterilization by thermal disinfection at 71$^\circ$C or Gamma sterilization \cite{MBengue2023}. Printing with flexible materials such as TPU requires slower motion of the extruder compared to rigid material. However, early attempts to design a hand with a rigid material such as PLA failed due to rapid plastic deformation in flexion. We make a distinction between a two-finger (\nn{2}) and a three or more finger ($n>2$) hands described next.

\subsubsection{\nn{2}, Two-finger hands} 
\begin{figure}
    \centering
    \includegraphics[width=\linewidth]{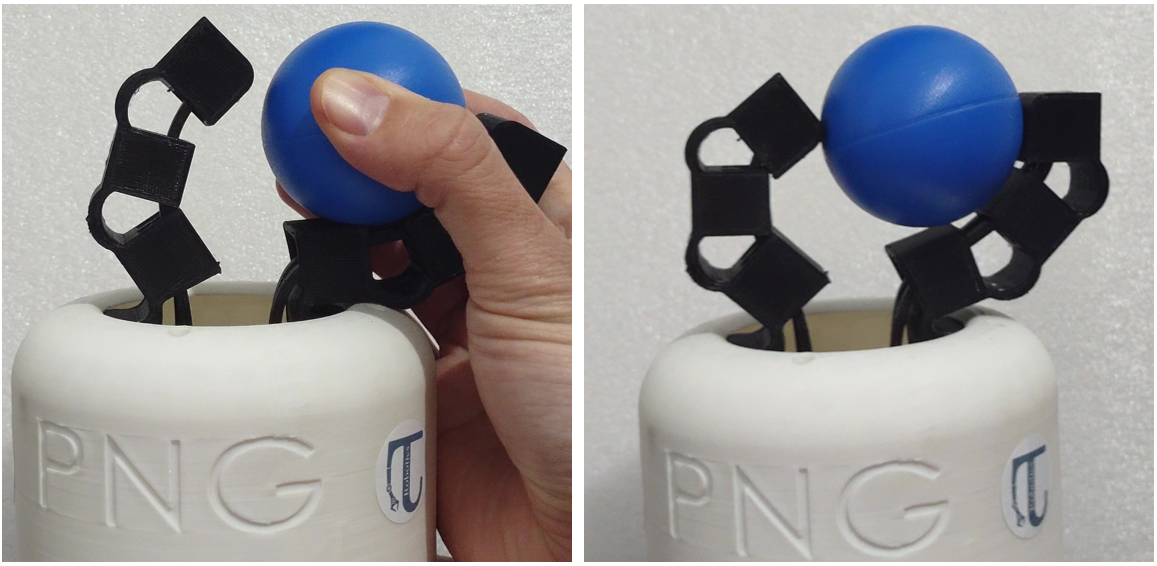}
    \caption{Demonstration of the differential in the PNG-2. In this case, (left) the right finger is manually constrained with a ball and the left finger compensates for the increase in tendon force $f_{in}$. (right) The ball is not centered in the resulted grasp.}
    \label{fig:differential}
\end{figure}

A \nn{2} hand is seen in Figure \ref{fig:main}b. The tendon bands of both fingers are connected at a pull loop. Later, a pulley from the universal base hooks onto the pull loop and pulled in order to achieve flexion of the fingers. The pulley functions also as a differential where it can roll along the bands and distribute the pulling force. In other words, the force is divided to each finger based on the reaction load it experiences. This is demonstrated in Figure \ref{fig:differential}. Since the \nn{2} hand is flat, it is 3D printed easily while laying horizontal on the heated bed as demonstrated in Figure \ref{fig:bed}a.

\subsubsection{Three or more finger hands}
While many fingers can be used, more than four fingers in discrete rotational symmetry do not offer much advantages and may also compromise performance due to self-collisions. Hence, we evaluate only \nn{3} and \nn{4} hands while configurations with more fingers are possible. The fingers of a hand are aligned in discrete rotational symmetry as seen in Figure \ref{fig:main}c. The bands of the fingers meet at a pull loop at the bottom center. In such form, a limited differential mechanism is provided where the pull loop can move with regards to the center axis of the hand upon different reaction loads on the fingers. The hand is 3D printed such that the fingers are spread across the heated bed with the support flanges at the top center. An example of a 3D printed \nn{4} outcome on the heated bed is seen in Figure \ref{fig:bed}b.

\begin{figure*}
    \centering
    \begin{tabular}{cc}
        \includegraphics[width=0.49\linewidth]{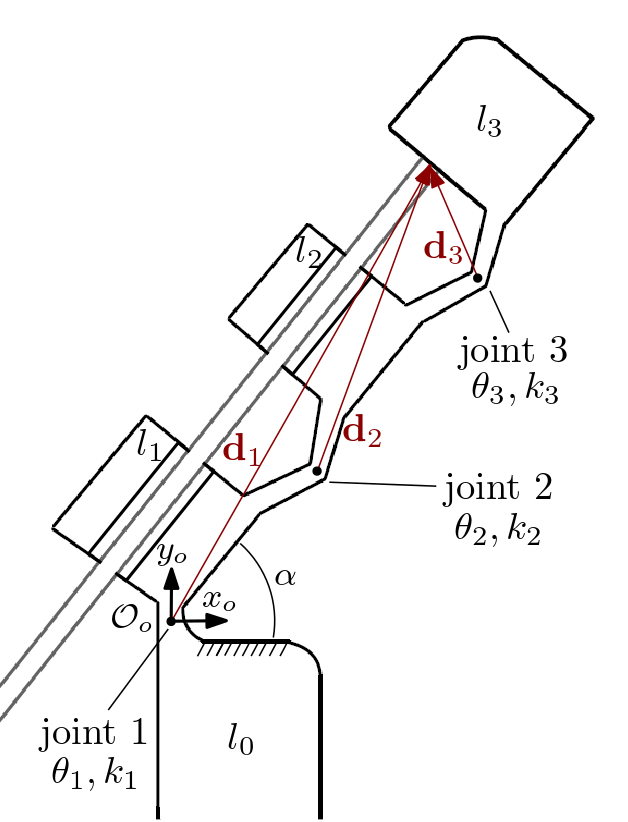} &  \includegraphics[width=0.39\linewidth]{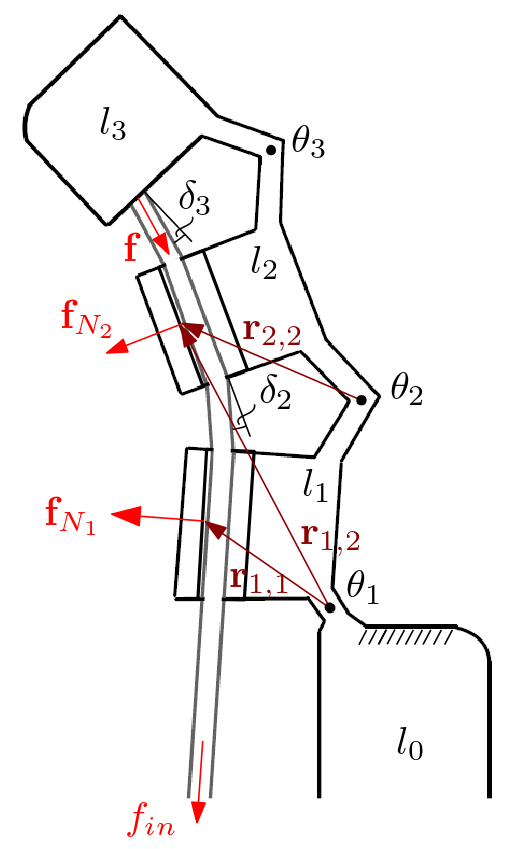}\\
         (a) &  (b) \\
    \end{tabular}
    \caption{Illustration of a PNG finger with $m=3$ joints when (a) relaxed and (b) in flexion.}
    \label{fig:theo_examp}
\end{figure*}
\subsection{Scaling of the hand}
\label{sec:scaling}

The unique formation and its one-shot property enable scaling of the hand to various sizes. Let $\kappa$ denote the scaling factor and define the reference hand where the length of the fingers is 52~mm, i.e., $\kappa=1$ when the finger is of 52~mm length. 
Hence, a practitioner can scale the CAD model of the reference hand by a desired factor $\kappa$ in order for the hand to fit a set of objects or tasks. In such scaling, all length parameters of the hand are scaled by $\kappa$ such that proportionality is maintained. Consequently, one can simply scale the CAD model in a CAD design software before printing, or in the slicer software of the 3D printer. A set of scaled \nn{2} hands can be seen in Figure \ref{fig:main}b.

\subsection{Universal Base}
\label{sec:base}

The universal base, illustrated in Figure \ref{fig:main}d, is a non-disposable part of the \n hand. The base consist of a shell, Dynamixel XH-540 actuator, pulley band and tightening cap. The lower part of the shell is the mounting plate for possible fixture to a robotic arm. Also, the actuator is fixed to the bottom of the inner shell with a set of screws. The pulley band is attached with screws on one end to the servo horn of the actuator while the other end connects to the pull loop of the 3D printed hand. The cylindrical shaped shell includes a thread on its round outer surface. Similarly, the tightening cap has an internal matching thread. Furthermore, four conical grooves are included on the top surface such that they radially slice the threaded part. The grooves continue from the top surface along the round surface. The grooves are designated for fixturing the mounting pins of the 3D printed hand. With their conical shape, varying sizes of scaled hands and mounting pins can be fixed to the base. Consequently, screwing the tightening cap on the thread applies pressure on the mounting pins as well as on the support flanges of the hand yielding a tight fix. The cap has an inner fillet that pushes the fingers to their exact pose.

With the proposed design of the universal base, it can accommodate scaled hands in the range $\kappa\in[1,2]$. In order to accommodate small scale hands, a reduction sleeve is added as seen in the right hand side of Figure \ref{fig:main}d. The  sleeve also has conical grooves and a smaller tightening cap for fixturing smaller scaled hands. Hence, it enables to accommodate hands of scale $\kappa\in[0.4,1.1]$. Note that the illustrations in Figure \ref{fig:main}, with and without the sleeve, show four grooves that can accommodate hands of two or four fingers. However, more grooves can be added in order to enable the use of three or more fingers. In our work, excluding the actuator and screws, the universal base is 3D printed. Nevertheless, one can fabricate the base in various designs including using metal components or sealed for liquid protection.

\section{Finger Model}
\label{sec:Modeling}

We now present a mechanical analysis of a single finger. 
Due to the elasticity of the 3D printed finger, the flexible joints can be considered as conventional joints with springs as seen in Figure \ref{fig:theo_examp}a. Hence, the system is analysed as a kinematic chain of rigid links connected by rotational joints with springs. Consider a general finger with $m+1$ links $l_0,l_1,\ldots,l_m$ and $m$ joints $\theta_1\ldots\theta_m$. Link $l_0$ is the support flange mounted to the universal base and, thus, considered immobilized. Joint $i$ is the angle between links $l_{i-1}$ and $l_i$. In addition, the flexibility of joint $\theta_i$ is represented by a torsional spring at the center of the joint with stiffness coefficient $k_i$. We assume quasi-static motions where dynamic effects and damping can be neglected. Also, we assume that joint angles $\theta_2\ldots\theta_m$ are equal to zero when no load is exerted on the band. This is true for fingers printed in this work while some initial curvature can be given in other designs. Joint $\theta_1$, on the other hand, is generally printed with an initial angle $\alpha$ such that $\theta_1=\alpha+\Delta\theta_1$ where $\Delta\theta_1$ is the angle change due to finger flexion. While the tendons are moving within the fingers, some friction effects may occur. However, we argue that friction is neglectable and validate this in experiments. 

Force $f_{in}$ is the tendon force exerted by the actuator on the band. The band runs through link $l_1,\ldots,l_{m-1}$ and is connected to the most distal one $l_m$. Assuming symmetry of middle links $l_1,\ldots,l_{m-1}$, it can be proven geometrically (see \cite{Wang2011}) that 
\begin{equation}
    \delta_i=\frac{\theta_i}{2}
\end{equation}
where $\delta_i$ is the angle between the band and link $i$ as seen in Figure \ref{fig:theo_examp}b. Force vector $\ve{f}\in\mathbb{R}^3$ exerted on link $l_m$ with respect to the finger coordinate frame $\mathcal{O}_o$ is given by
\begin{equation}
    \ve{f} = f_{in}R_z\left(\sum_{j=1}^m\theta_j\right)\begin{pmatrix}-\sin\delta_m\\-\cos\delta_m\\0\end{pmatrix}.
\end{equation}
where $R_z(\cdot)\in SO(3)$ is the rotation matrix about the $z$-axis. In the case where the finger is printed with an initial curvature, i.e., $\theta_i\neq0$ when $f_{in}=0$, then we consider the angle change $\Delta\theta_i$ instead of the absolute angle $\theta_i$. In addition to the force on $l_m$, force $f_{N_i}\in\mathbb{R}^3$ is exerted on link $l_i$ by the bands and is given by
\begin{equation}
    \ve{f}_{N_i} = f_{in}R_z\left(\sum_{j=1}^i\theta_j\right)\begin{pmatrix}0\\-\sin\delta_i\\0\end{pmatrix},
\end{equation}
for any $i<m$.

Define $\ve{d}_i\in\mathbb{R}^3$ for $i<m$ as the vector from joint $\theta_i$ to the center of contact between link $l_m$ and the band. Similarly, 
vector $\ve{r}_{j,i}\in\mathbb{R}^3$ is the vector from joint $j$ to the center of contact between the band and link $i$. Vectors $\ve{d}_i$ and $\ve{r}_{j,i}$ can be computed by simple forward kinematics of the finger. Also, let 
\begin{equation}
    \label{eq:sigma_spring}
    \sigma_i=\begin{pmatrix}
        0 & 0 & k_i\theta_i
    \end{pmatrix}^T
\end{equation}
be the torque on joint $i$ resulting from torsional spring deflection. From force analysis on the distal link $l_m$, the torque on joint $\theta_m$ is given by
\begin{equation}
    \label{eq:tau_m}
    \tau_m=\ve{f}\times\ve{d}_m-\sigma_m.
\end{equation} 
When $i<m$, the joint torques are
\begin{equation}
    \label{eq:tau_i}
    \tau_i = \ve{f}\times\ve{d}_i + \sum_{j=i}^{m-1} (\ve{f}_{N_i}\times\ve{r}_{j,i}) -\sigma_i
\end{equation}

As discussed above, the motion of the finger is considered quasi-static. Hence, the system is in torque equilibrium yielding $\tau_i=0$ for all $i=1,\ldots,m$. Consequently, the following result can be derived from \eqref{eq:sigma_spring}-\eqref{eq:tau_i}:
\begin{equation}
    \label{eq:model1}
    \theta_i=\frac{\|\ve{u}_i\|}{k_i}  
\end{equation}
where 
\begin{equation}
    \label{eq:model2}
    \ve{u}_i=\begin{cases}
       \ve{f}\times\ve{d}_i + \sum_{j=i}^{m-1} (\ve{f}_{N_i}\times\ve{r}_{j,i}), &\quad i<m. \\ 
       \ve{f}\times\ve{d}_m, &\quad i=m\\
     \end{cases}
\end{equation}
Given stiffness coefficients $k_1,\ldots,k_m$, model \eqref{eq:model1}-\eqref{eq:model2} can provide a mapping between the band pulling force to joint angles. In other words, one can optimize the number of joints and geometric values in order to acquire a desired finger flexion. The model will be validated through experiments in the next section.

\section{Experiments}
\label{sec:experiments}

In the following experiments, we validate the finger model and test the capabilities of various \n hands with different configurations. As described previously, all hands were 3D printed on a conventional FDM printer with a TPU filament. 
A video of the experiments and demonstrations is available on YouTube\footnote{Video link: \url{https://youtu.be/dk5-teuzLGE}}.

\begin{figure}[h]
    \centering
    \includegraphics[width=\linewidth]{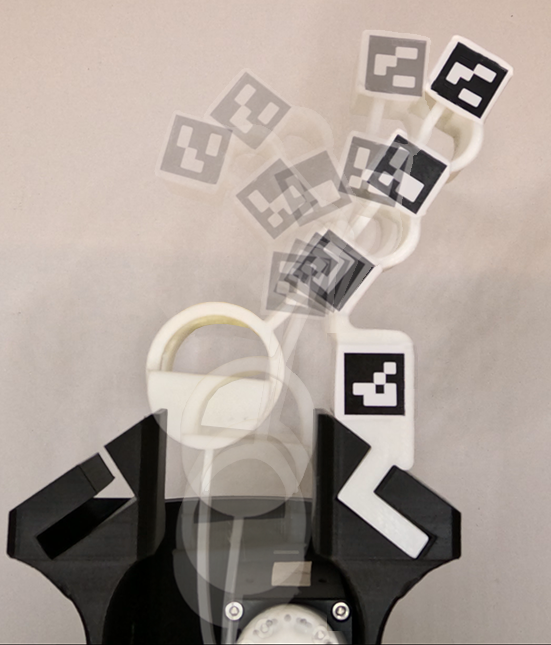}
    \caption{Experimental setup for verification of the finger model with scale $\kappa=1.5$. Fiducial markers on the finger are observed by a camera. Flexion of the finger is seen during the pull of the tendon by the actuator at the bottom.}
    \label{fig:Bends_exp}
\end{figure}

\subsection{Finger Model Verification}

The finger model proposed in Section \ref{sec:Modeling} maps input forces to finger flexions. In this section, the model is experimentally verified. An experimental setup, seen in Figure \ref{fig:Bends_exp}, was designed where fiducial markers are positioned on the links of a finger with $m=3$ links and initial angle $\alpha=50\degree$. Tracking the position and orientation of the markers with a camera enables to compute an estimation of joint angles $\tve{q}=(\tilde{\theta}_1, \tilde{\theta}_2, \tilde{\theta}_3)^T$. In addition, the pulling force $f_{in}$ is computed with the pulley radius and actuator internal sensing of torque. 

Data was collected by recording 16 flexions of a finger with scale $\kappa=1.5$ yielding $N=160$ samples. Model \eqref{eq:model1}-\eqref{eq:model2} can be formulated as 
\begin{equation}
    f_\ve{k}(f_{{in}},\tve{q})=0
\end{equation}
where $\ve{k}=(k_1, k_2, k_3)$ is the vector of unknown torsional stiffness coefficients. Hence, estimated coefficients vector $\ve{k}^*$ can be found using the collected data and solving the following minimization problem 
\begin{equation}
    \label{eq:min}
    \ve{k}^*=\underset{\ve{k}}{\arg\min} \sum_{i=1}^N f_\ve{k}(f_{{in}_i},\tve{q}_i).
\end{equation}
Problem \eqref{eq:min} is solved using the BFGS algorithm \cite{fletcher1987}. 

\begin{figure}
    \centering
    \includegraphics[width=\linewidth]{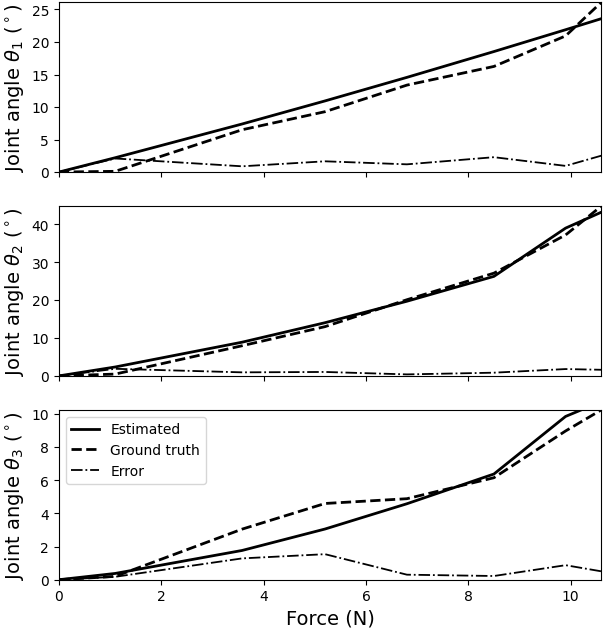}
    \caption{Estimated, ground-truth and estimation error of joint angles $\theta_1$, $\theta_2$ and $\theta_3$ during flexion with regards to pulling force $f_{in}$.}
    \label{fig:theo_theta}
\end{figure}
\begin{figure}
    \centering
    \includegraphics[width=\linewidth]{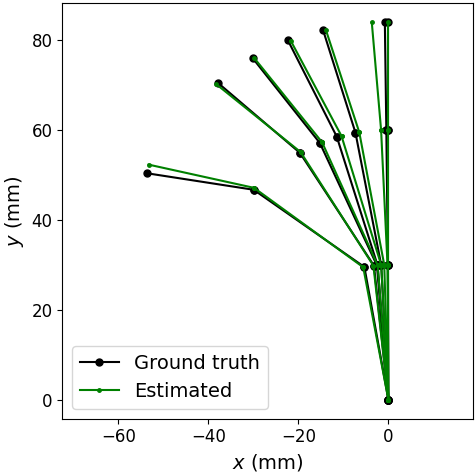}
    \caption{Finger flexion comparison of the measured ground-truth versus an estimation with model \eqref{eq:model1}-\eqref{eq:model2}.}
    \label{fig:theo_test}
\end{figure}

The solution of \eqref{eq:min} yielded joint stiffness values $k_1=28.48\frac{Nm}{rad}$ and $k_2=k_3=4.05\frac{Nm}{rad}$. By re-applying these values to \eqref{eq:model1}-\eqref{eq:model2}, one can estimate the motion of the finger. Figure \ref{fig:theo_theta} presents such an estimation over additional test data of recorded finger flexions. Results show the measured angles versus the estimated ones along with the estimation error. The mean angle error over all finger configurations in the data is $1.04\degree\pm0.74\degree$. A slightly large error can be seen at the early beginning of the $\theta_1$ motion due to the unmodeled tightening phase of the band from a loose state. Figure \ref{fig:theo_test} illustrates the measured and estimated fingers. While some assumptions were made in formulating the model, the results show that these are valid and the error is low. Hence, a practitioner can use the proposed model in order to optimize the parameters of the finger to fit various tasks. Also, the model can be used for motion control and for training a policy in a reinforcement learning scheme \cite{Sintov2019,Azulay2023}.



\subsection{Fabrication and Assembly Time}

\begin{figure}
    \centering
    \includegraphics[width=\linewidth]{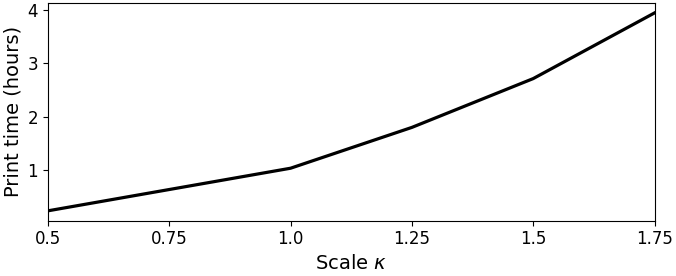}
    \caption{3D printing time of a \nn{2} hand with regards to scale $\kappa$.}
    \label{fig:print_time}
\end{figure}

\begin{figure*}
    \centering
    \includegraphics[width=\linewidth]{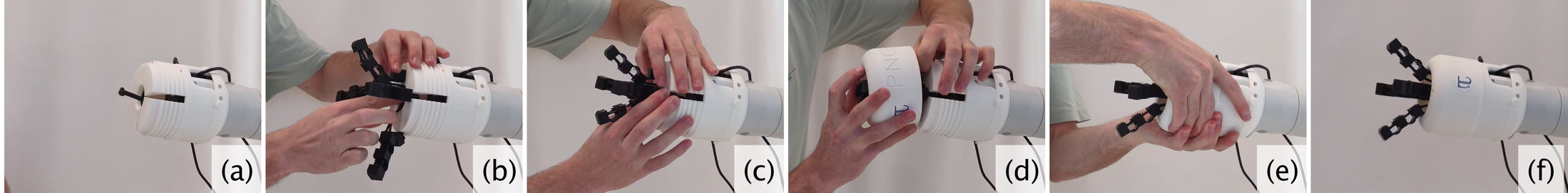}
    \caption{Mounting a \nn{4} hand to the base: (a) The tightening cap is removed and the base is ready for mounting a hand with the pulley band sticking out; (b)-(c) The 3D printed hand is mounted by connecting the pulley band to the pull loop and the mounting pins to the base grooves; (d) The fingers pass through the tightening cap; (e) The tightening cap is screwed onto the base; (f) The hand is ready for operation.}
    \label{fig:assembly_PNG4}
\end{figure*}

Experiments are now conducted in order to estimate the time from initiating the 3D printing to when the \n hand is assembled on the universal base and ready to grasp. Figure \ref{fig:print_time} shows the printing time of \nn{2} hands on a Bambu Lab X1-Carbon 3D Printer with regards to the scale $\kappa$. After printing, a cleaning step  is performed where the \n hand is removed from the heated bed and the support is pealed off. In the final step, the hand is assembled on the universal base as seen in Figure \ref{fig:assembly_PNG4}. Several novice users were asked to perform the cleaning and assembly steps after brief instructions. Time measurements for the cleaning yielded 62 seconds in average. The assembly of the hand on the base took an average of 65 seconds for the first tries. However, after several attempts, the users were able to assemble the hand in 25 seconds. Overall, the results show that the \n hand is printed relatively fast and the quickest, compared to state-of-the-art, to reach from the 3D printer to actual operation. 

\begin{figure}
    \centering
    \includegraphics[width=\linewidth]{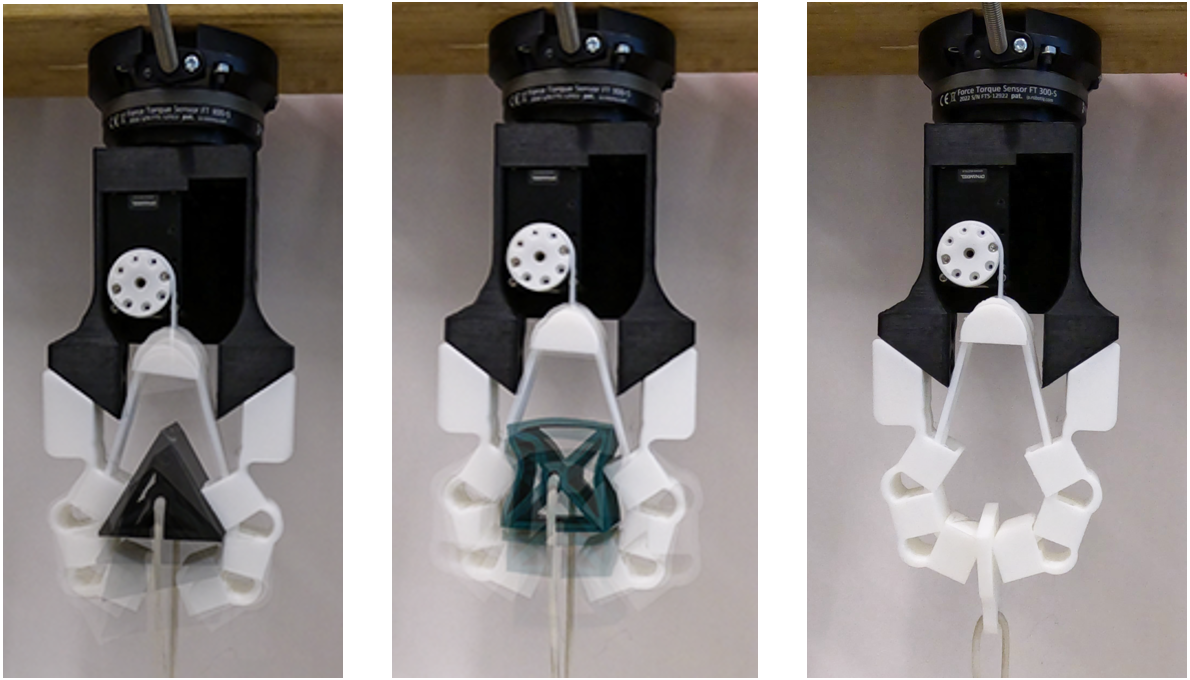}
    \caption{Experimental setup for measuring the maximum load a \nn{2} hand can withhold. The object was pulled downward while measuring the force exerted by the actuator in order to maintain equilibrium. The objects seen are (from left to right) rigid triangular prism, elastic cuboid and thin sheet.}
    \label{fig:exp_setup}
\end{figure}
\begin{figure}
    \centering
    \includegraphics[width=\linewidth]{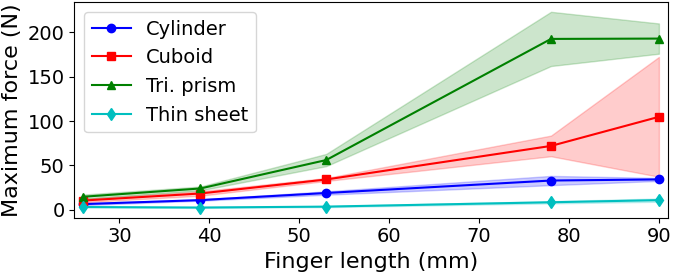}
    \caption{Maximum load for \n{2} hands printed with TPU-95A and grasping a rigid object, with regards to finger length (corresponding to scale).}
    \label{fig:max_load_95A}
\end{figure}
\begin{figure}
    \centering
    \includegraphics[width=\linewidth]{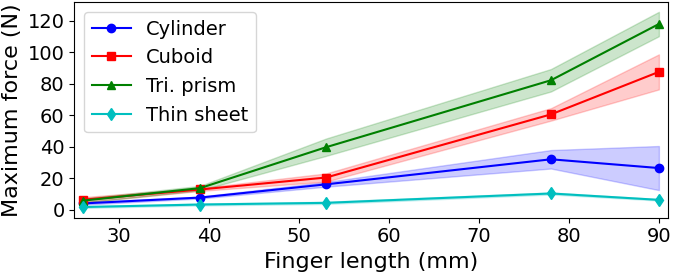}
    \caption{Maximum load for \n{2} hands printed with TPU-85A and grasping a rigid object, with regards to finger length (corresponding to scale).}
    \label{fig:max_load_85A}
\end{figure}
\begin{figure}
    \centering
    \includegraphics[width=\linewidth]{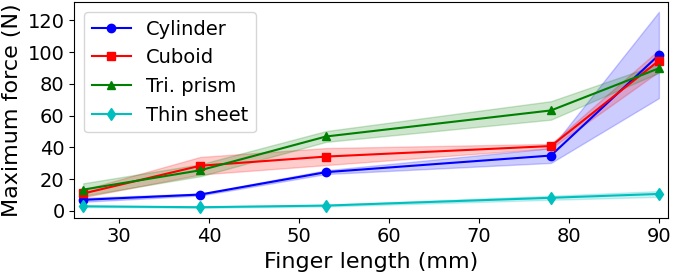}
    \caption{Maximum load for \n{2} hands printed with TPU-95A and grasping a flexible object, with regards to finger length (corresponding to scale).}
    \label{fig:max_load_95A_flex}
\end{figure}
\begin{figure}
    \centering
    \includegraphics[width=\linewidth]{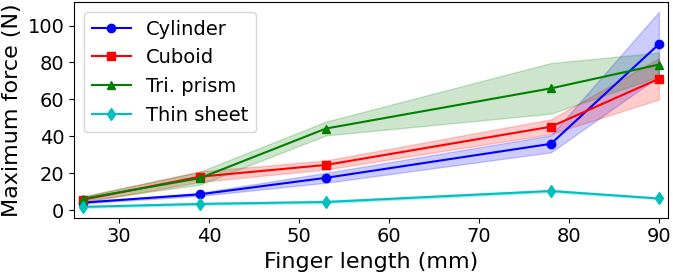}
    \caption{Maximum load for \n{2} hands printed with TPU-85A and grasping a flexible object, with regards to finger length (corresponding to scale).}
    \label{fig:max_load_85A_flex}
\end{figure}
\begin{figure*}[]
    \centering
    \begin{minipage}{.52\textwidth}

    \centering
    \caption{Success rates of picking-up and relocating various objects with \nn{2} hands of different scale}
    \label{tb:scale_sr}
    \begin{adjustbox}{width=\linewidth}
    \begin{tabular}{l|cccccc}
    \hline
    \multirow{2}{*}{\diagbox{Object}{Scale}} & $\kappa=0.5$ & $\kappa=0.75$ & $\kappa=1$ & $\kappa=1.5$ & $\kappa=1.75$ & $\kappa=1.75$\\
    & (\%) & (\%) & (\%) & (\%) & (\%) & w/ nails (\%)\\\hline
    
    Hammer              & 0 & 0  & 60 & \cellcolor[HTML]{C0C0C0}100 & \cellcolor[HTML]{C0C0C0}100 & \cellcolor[HTML]{C0C0C0}100 \\
    
    Bottle (0.5L)       & 0 & 70  & 90  & \cellcolor[HTML]{C0C0C0}100 & \cellcolor[HTML]{C0C0C0}100 & \cellcolor[HTML]{C0C0C0}100 \\
    
    Water glass & 0 & 0     & 0  & 70 & \cellcolor[HTML]{C0C0C0}100 & \cellcolor[HTML]{C0C0C0}100 \\
    
    Chewing gum pack   & 60 & 80 & \cellcolor[HTML]{C0C0C0}100 & \cellcolor[HTML]{C0C0C0}100 & \cellcolor[HTML]{C0C0C0}100 & \cellcolor[HTML]{C0C0C0}100 \\
    
    Cheetos bag         & 10 & 50 & \cellcolor[HTML]{C0C0C0}100 & \cellcolor[HTML]{C0C0C0}100 & \cellcolor[HTML]{C0C0C0}100 & \cellcolor[HTML]{C0C0C0}100 \\
    
    Rubber duck         & 70 & 90  & \cellcolor[HTML]{C0C0C0}100 & \cellcolor[HTML]{C0C0C0}100 & \cellcolor[HTML]{C0C0C0}100 & \cellcolor[HTML]{C0C0C0}100 \\
    
    USB flash drive     & \cellcolor[HTML]{C0C0C0}100 & 80 & 80 & 90 & 80 & \cellcolor[HTML]{C0C0C0}100  \\
    
    Toilet paper roll   & \cellcolor[HTML]{C0C0C0}100 & \cellcolor[HTML]{C0C0C0}100 & \cellcolor[HTML]{C0C0C0}100 & \cellcolor[HTML]{C0C0C0}100 & \cellcolor[HTML]{C0C0C0}100 & \cellcolor[HTML]{C0C0C0}100 \\
    
    Same gripper      & \cellcolor[HTML]{C0C0C0}100 & \cellcolor[HTML]{C0C0C0}100 & \cellcolor[HTML]{C0C0C0}90 & 80 & 60 & 80 \\
    
    Bunch of keys    & \cellcolor[HTML]{C0C0C0}100 & \cellcolor[HTML]{C0C0C0}100 & 90 & 80 & 70 & \cellcolor[HTML]{C0C0C0}100 \\ 
    
    Wet wipes & \cellcolor[HTML]{C0C0C0}100 & \cellcolor[HTML]{C0C0C0}100 & \cellcolor[HTML]{C0C0C0}100 & \cellcolor[HTML]{C0C0C0}100 & \cellcolor[HTML]{C0C0C0}100 & \cellcolor[HTML]{C0C0C0}100 \\ 
    
    Coin                & 0 & 20 & 20 & 40 & 10 & \cellcolor[HTML]{C0C0C0}100 \\ \hline
    \end{tabular}
    \end{adjustbox}
    \end{minipage}%
    \hfill
    \begin{minipage}{0.46\textwidth}

    \centering
    \caption{Success rates of picking-up and relocating various objects with two-, three- and four-finger hands}
    \label{tb:fingerNum_sr}
    \begin{adjustbox}{width=\linewidth}
    \begin{tabular}{l|ccc|ccc}
    \hline
    \multirow{3}{*}{\diagbox{Object}{Hand}}  & \multicolumn{3}{c|}{$\kappa=0.75$} & \multicolumn{3}{c}{$\kappa=1.25$} \\
     & \nn{2} & \nn{3} & \nn{4} & \nn{2} & \nn{3} & \nn{4} \\
    & (\%) & (\%) & (\%) & (\%) & (\%) & (\%) \\\hline
    Hammer              & 0 & 0 & 0 &  \cellcolor[HTML]{C0C0C0}100   &  \cellcolor[HTML]{C0C0C0}100   &  \cellcolor[HTML]{C0C0C0}100   \\
    Bottle (0.5L)       & 70 & 0 & \cellcolor[HTML]{C0C0C0}100 & \cellcolor[HTML]{C0C0C0}100 & 90    &  \cellcolor[HTML]{C0C0C0}100   \\
    Water glass         & 0 & 0 & 50 & 70    &  \cellcolor[HTML]{C0C0C0}100   &  \cellcolor[HTML]{C0C0C0}100   \\
    Chewing gum pack   & 80 & 90 & \cellcolor[HTML]{C0C0C0}100 &  \cellcolor[HTML]{C0C0C0}100   &  \cellcolor[HTML]{C0C0C0}100   &  \cellcolor[HTML]{C0C0C0}100   \\
    Cheetos bag         & 50 & 50 & 80 &  \cellcolor[HTML]{C0C0C0}100   &  \cellcolor[HTML]{C0C0C0}100   &  \cellcolor[HTML]{C0C0C0}100   \\
    Rubber duck         & 90 & \cellcolor[HTML]{C0C0C0}100 & \cellcolor[HTML]{C0C0C0}100 &  \cellcolor[HTML]{C0C0C0}100   &  \cellcolor[HTML]{C0C0C0}100   &  \cellcolor[HTML]{C0C0C0}100   \\
    USB flash drive     & 80 & 80 & 90 & \cellcolor[HTML]{C0C0C0}90    & 10    & 40    \\
    Toilet paper roll   & \cellcolor[HTML]{C0C0C0}100 & \cellcolor[HTML]{C0C0C0}100 & \cellcolor[HTML]{C0C0C0}100 &  \cellcolor[HTML]{C0C0C0}100   &  \cellcolor[HTML]{C0C0C0}100   &  \cellcolor[HTML]{C0C0C0}100   \\
    Same gripper        & \cellcolor[HTML]{C0C0C0}100 & \cellcolor[HTML]{C0C0C0}100 & \cellcolor[HTML]{C0C0C0}100 & 80    &  \cellcolor[HTML]{C0C0C0}100   &  \cellcolor[HTML]{C0C0C0}100   \\
    Bunch of keys       & \cellcolor[HTML]{C0C0C0}100 & \cellcolor[HTML]{C0C0C0}100 & \cellcolor[HTML]{C0C0C0}100 & 80    &  \cellcolor[HTML]{C0C0C0}100   &  \cellcolor[HTML]{C0C0C0}100   \\ 
    Wet wipes & \cellcolor[HTML]{C0C0C0}100 & \cellcolor[HTML]{C0C0C0}100 & \cellcolor[HTML]{C0C0C0}100 &  \cellcolor[HTML]{C0C0C0}100   &  \cellcolor[HTML]{C0C0C0}100   &  \cellcolor[HTML]{C0C0C0}100   \\ 
    Coin                & 20 & 0 & 0 & \cellcolor[HTML]{C0C0C0}40    & 0     & 0     \\ \hline
    \end{tabular}
    \end{adjustbox}
    \end{minipage}
\end{figure*}

\begin{figure*}[h]
    \centering
    \includegraphics[width=\linewidth]{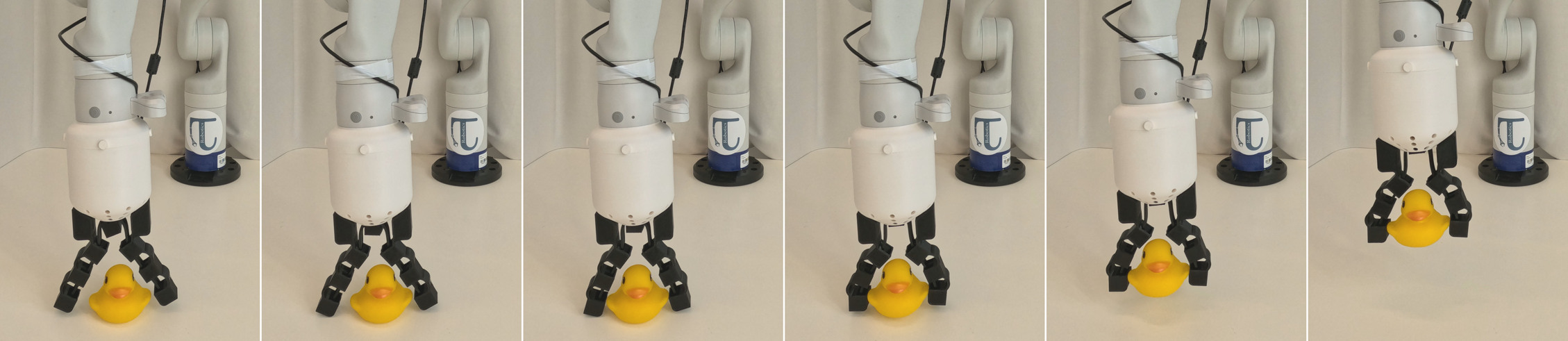}
    \caption{Pick-up experiment of a rubber duck with a \nn{2} hand of $\kappa=1.5$.}
    \label{fig:sn_duck}
\end{figure*}
\begin{figure*}[h]
    \centering
    \includegraphics[width=\linewidth]{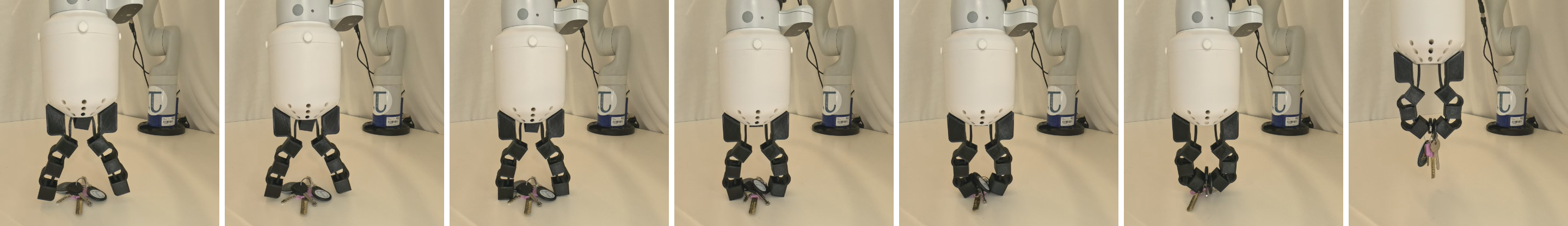}
    \caption{Pick-up experiment of keys with a \nn{2} hand of $\kappa=1.5$.}
    \label{fig:sn_keys}
\end{figure*}
\begin{figure*}
    \centering
    \includegraphics[width=\linewidth]{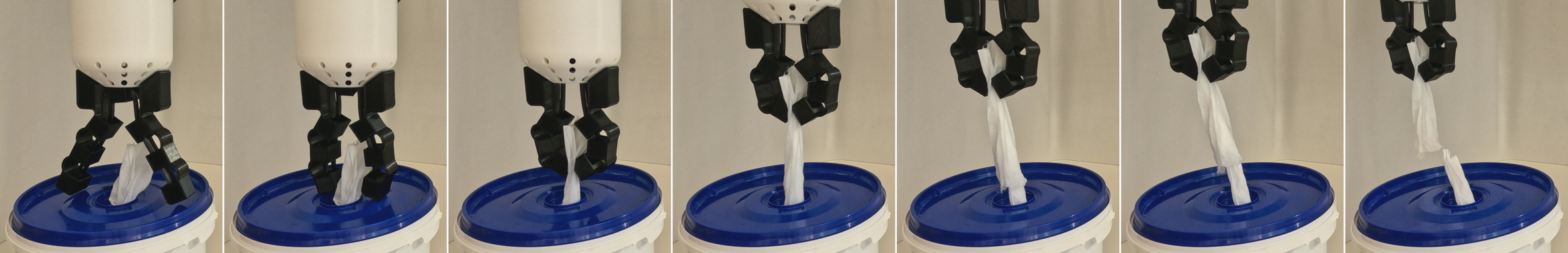}
    \caption{Pick-up experiment of a wet-wipe with a \nn{2} hand of $\kappa=1.75$.}
    \label{fig:sn_wet}
\end{figure*}
\begin{figure*}[h]
    \centering
    \includegraphics[width=\linewidth]{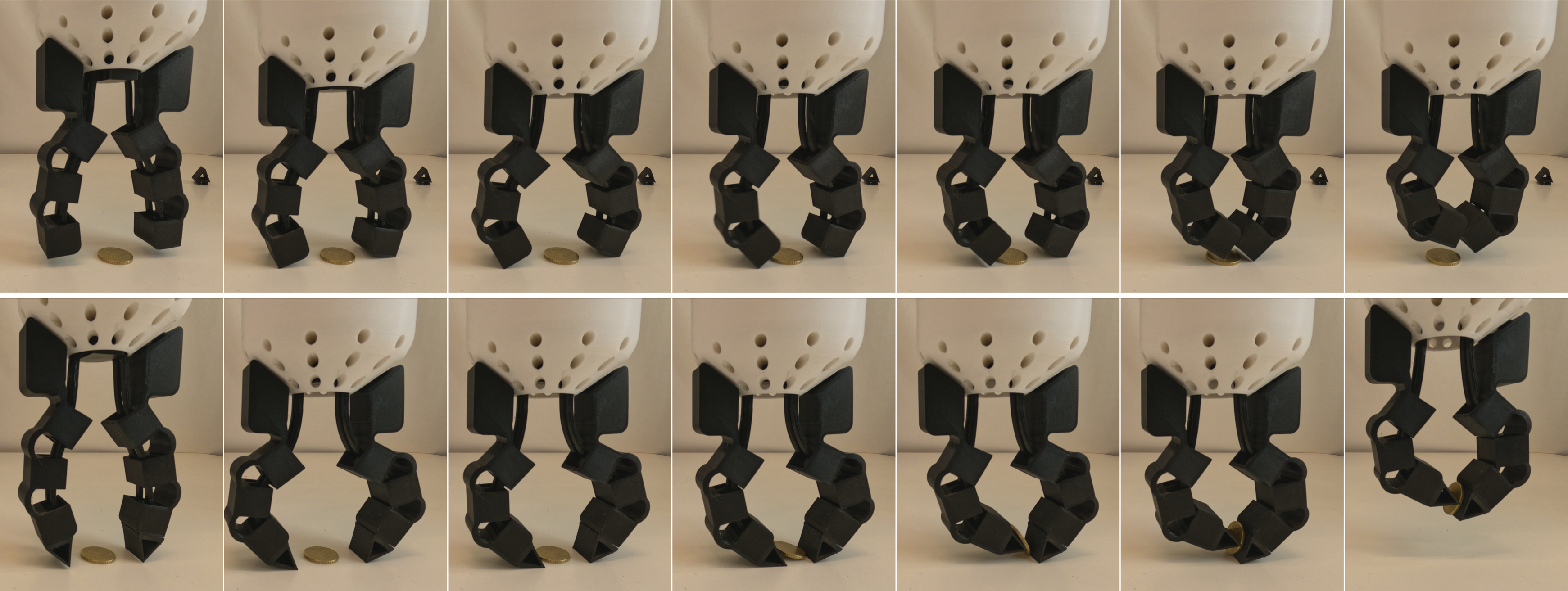}
    \caption{Pick-up experiment of a coin with a \nn{2} hand of $\kappa=1.75$ (top) not-equipped and (bottom) equipped with fingernails. The fingernails enable to flip the coin and grasp it on the flat sides.}
    \label{fig:sn_coin}
\end{figure*}


\subsection{Maximum load}
\label{sec:max_load}

We wish to analyze the maximum load that a two-finger gripper can bare. Hence, we have built an experimental setup, seen in Figure \ref{fig:exp_setup}, where grippers of various sizes can be tested. The setup consists of an early design of the universal base with a Robotiq FT-300 Force/Torque (F/T) sensor fixed vertically on a rigid frame. A gripper mounted on the universal base grasps an object that is pulled downwards by a string. The pulling force is increased and recorded until reaching a maximum when the object slips. Each experiment was averaged over four repeated iterations while placing the object in arbitrary positions and orientation angles within the hand.


We test five gripper sizes of finger lengths 25.6mm, 38.5mm, 52mm, 77.2mm and 89.7mm, which correspond to scale $\kappa=0.5$, $\kappa=0.75$, $\kappa=1$, $\kappa=1.5$ and $\kappa=1.75$ of the reference size. The hands are seen in Figure \ref{fig:main}d. Furthermore, two printing materials, TPU-95A and TPU-85A, are compared. Moreover, four shapes are used: cylinder, cuboid, triangular prism and a thin sheet for pinch grasping. However, a set of objects was printed for each gripper with corresponding sizes in order to match its scale. In addition, we test rigid (PLA) and flexible (TPU) objects. 

Figures \ref{fig:max_load_95A} and \ref{fig:max_load_85A} present the mean maximum loads for grasping rigid objects and for TPU-95A and TPU-85A, respectively. Similarly, Figures \ref{fig:max_load_95A_flex} and \ref{fig:max_load_85A_flex} present the mean maximum loads for grasping flexible objects and for TPU-95A and TPU-85A, respectively. The results show that the grippers can withstand significantly high loads roughly proportional to size. The highest load of up to 200~N was achieved by the largest gripper and with rigid objects. When comparing between TPU-95A and TPU-85A, the former can hold higher forces. TPU-95A is much less flexible having a Young modulus of 26 MPa compared to TPU-85A with 12MPa. Also, rigid objects are easier to grasp as the flexible object deform between the fingers and slip. While triangular prisms are best grasped, pinch grasps are harder to firmly maintain due to slips.

\subsection{Mechanical characteristics  analysis}



We next perform a fatigue analysis and test the ability of the PNG hand to withstand cyclic loading. A newly printed PNG-2 hand with $\kappa=1.5$ was set to repeatedly grasp a drill handle of diameter 45~mm. Within each grasp, the hand was closed tight with a tendon force of approximately $f_{in}\approx75N$. Following every 1,000 trial cycles, the hand was disassembled and the structural integrity was visually inspected. In addition to visual inspection, the hand was tested in loading as performed in Section \ref{sec:max_load}. The experiment was run for 10,000 cycles. No plastic deformations or any damages were observed but only some wear on the distal contact pads. In addition, loading tests yielded similar results to the ones presented in Figure \ref{fig:max_load_95A}. Overall, the PNG hand is validated to withstand substantial cyclic loading. 

In addition to fatigue testing, another experiment was conducted to test the durability of the hand. Different hands were strongly hit with an hammer multiple times followed by grasping experiments. Similarly, the hand was pushed by the robot toward the table and dragged. No damage was caused to the hands and they operated successfully. These experiments can be seen in the supplementary videos.

\subsection{Grasping Experiments}

\begin{figure*}[!htb]
    \centering
    \begin{minipage}{.48\textwidth}
        \centering
        \includegraphics[width=\linewidth]{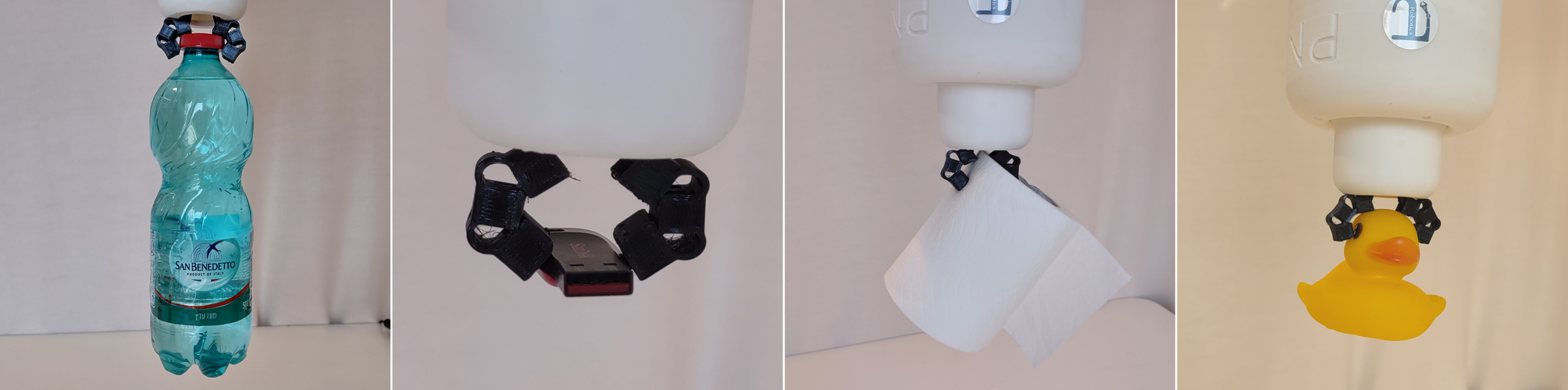}
        \caption{Snapshots of pick-up experiments of (left to right) bottle, keys, toilet paper roll and rubber duck with a \nn{2} hand of $\kappa=0.75$.}
        \label{fig:PNG2.75_objects}
    \end{minipage}%
    \hfill
    \begin{minipage}{0.48\textwidth}
        \centering
        \includegraphics[width=\linewidth]{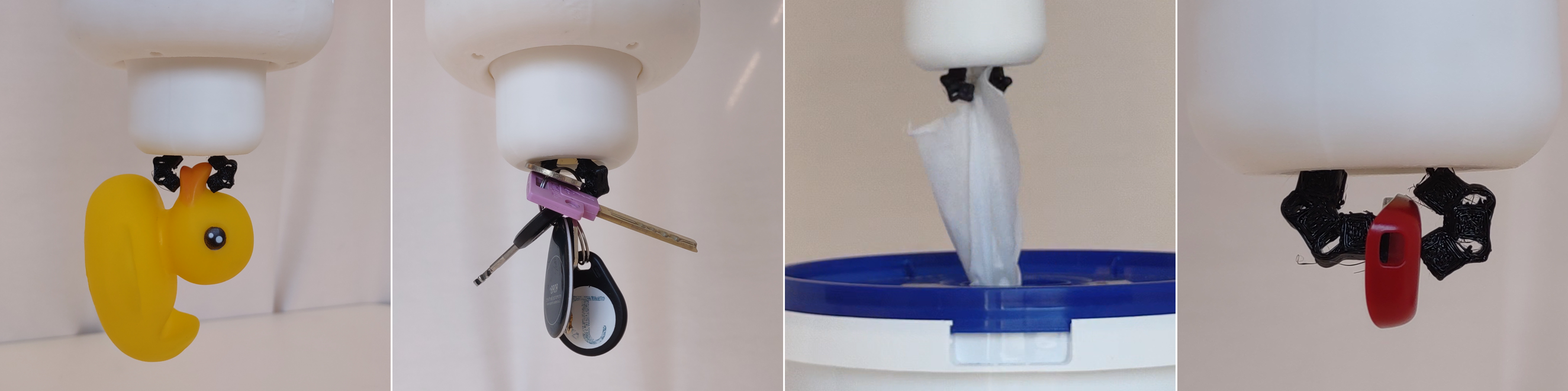}
        \caption{Snapshots of pick-up experiments of (left to right) rubber duck, keys, wet wipe and flash drive with a \nn{2} hand of $\kappa=0.5$.}
        \label{fig:PNG2.5_objects}
    \end{minipage}
\end{figure*}

In the next experiment, we evaluate the ability of PNG hands of various scales printed with TPU-95A to pick-up everyday objects. Note that some of the experiments in this section were not performed with the universal base presented in Section \ref{sec:base} but with earlier prototypes. Nevertheless, the tested \n hands are in their final form as proposed in this paper. We start by experimenting with five \nn{2} hands having a scale of $\kappa=0.5$, $\kappa=0.75$, $\kappa=1$, $\kappa=1.5$ and $\kappa=1.75$. In addition, we include trials with a \nn{2} hand of scale $\kappa=1.75$ having also fingernails. Each hand was mounted on the universal base while the former is fixed on the Kinova Gen3 robot arm. Various everyday objects were placed on a table for the robot to pick up. The robot was lowered down to the object and stopped at some arbitrary height in the vicinity of the object. After closing the PNG hand, the hand was pulled back up. The success rate for lifting each object over 10 trials is reported in Table \ref{tb:scale_sr}. The results show that smaller hands have more difficulties picking-up objects with large circumference such as the bottle or water glass. With larger hands, higher success rates are reached. Figures \ref{fig:sn_duck}-\ref{fig:sn_wet} show snapshots of pick-up trials with a \nn{2} hand of a rubber duck, bunch of keys and wet-wipes, respectively. Figure \ref{fig:sn_coin} shows a pick-up of a coin with and without the usage of finger-nails. The addition of fingernails evidently assisted in improving success rate in smaller objects such as the keys and coin. Figures \ref{fig:PNG2.75_objects} and \ref{fig:PNG2.5_objects} depict the picking-up of various objects with smaller \nn{2} hands of scale $\kappa=0.75$ and $\kappa=0.5$, respectively.

\begin{figure*}[h]
    \centering
    \includegraphics[width=\linewidth]{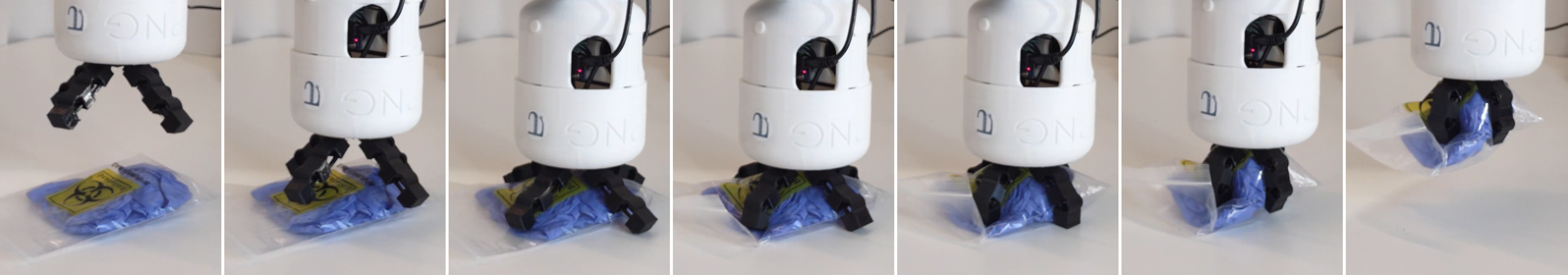}
    \caption{Pick-up experiment of a bio-hazard bag with a \nn{4} hand of $\kappa=1.5$.}
    \label{fig:sn_bio}
\end{figure*}

\begin{figure*}[!htb]
    \centering
    \begin{minipage}{.48\textwidth}
        \centering
        \includegraphics[width=\linewidth]{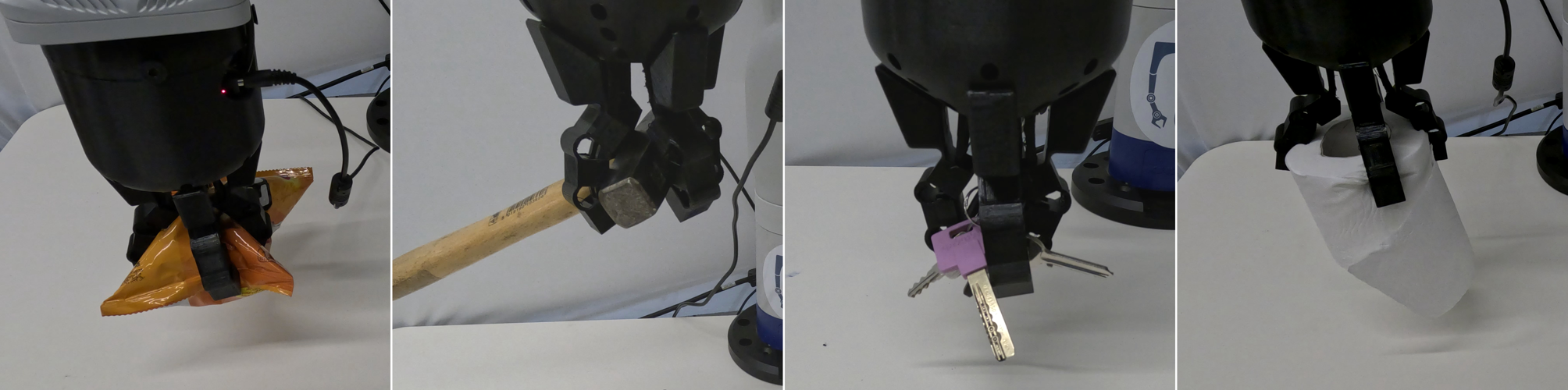}
        \caption{Snapshots of pick-up experiments of (left to right) Cheetos bad, hammer, keys and toilet paper roll with a \nn{4} hand of $\kappa=1$.}
        \label{fig:PNG4_objects}
    \end{minipage}%
    \hfill
    \begin{minipage}{0.48\textwidth}
        \centering
        \includegraphics[width=\linewidth]{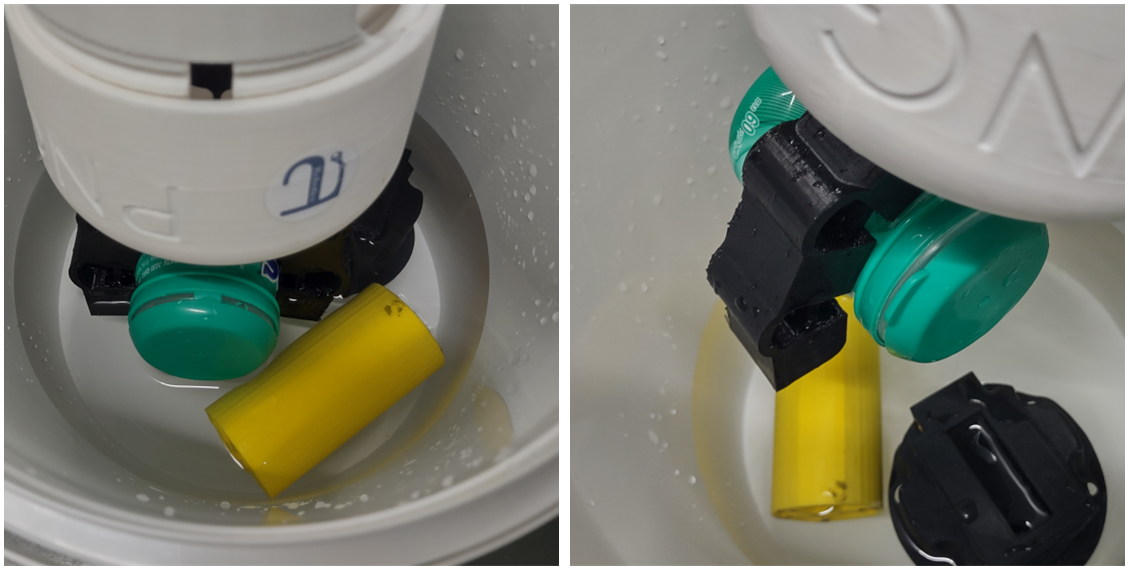}
        \caption{Snapshots of pick-up experiments of a chewing gum pack from a bucket of water  with a PNG-2 hand of $\kappa=1.75$.}
        \label{fig:liquid}
    \end{minipage}
\end{figure*}

The same experiments are now presented for \nn{3} and \nn{4} hands of scale $\kappa=0.75$ and $\kappa=1.25$. Table \ref{tb:fingerNum_sr} reports the picking-up success rate for all the hands over all objects and compared to \nn{2}. Results show that more fingers slightly improves success rate in general and smaller objects in particular. Nevertheless, a coin is almost impossible to pick-up without fingernails. Figure \ref{fig:sn_bio} and \ref{fig:PNG4_objects} show snapshot of picking-up a small bio-hazard bag and various other objects, respectively, with a \nn{4} hand of scale $\kappa=1$. 
Furthermore, and to demonstrate the ability of the PNG hand to work in complex environments, we also test the grasping of a chewing gum pack in a bucket of liquid with additional objects as seen in Figure \ref{fig:liquid}. The success rate out of ten attempts is 90\% with one slippage. In addition, the hand was able to further work in other tasks after getting wet. In summary, the results show high success rate for picking up various objects of different sizes with different hands. Moreover, it is possible to use hands of one scale (e.g., $\kappa=1.25$) for successfully picking up a large number of object classes.




\section{Conclusions}

In this paper, a novel robotic hand termed \textit{Print-n-Grip} (\nm) was presented. The \n hand is fabricated through one-shot 3D printing and, therefore, accessible and low-cost. The design of a finger includes built-in tendons removing the need for tedious string threading. Once printed, the hand can easily and rapidly be mounted on a universal base. Since it is easily fabricated, different 3D printed configurations of a \n hand can be mounted on the same universal base. A hand can be 3D printed with two or more fingers based on desired tasks or objects to grasp. Furthermore, the hand is scalable such that a practitioner can 3D print small or large \n hands simply by determining a scaling factor depending on the task or objects to be grasped. A model of a finger was proposed such that the pulling force of the tendon is mapped to flexion of the finger. The model was validated through experiments. 
With the model, one can design and optimize various new configurations of a finger and hand. In addition, a set of experiments presented in this paper has shown the ability of \n hands with different scales to lift heavy objects and successfully pick-up objects of different size and geometry.

The proposed \n hand enables a low-cost and disposable solution for chemical and medical applications. Once contaminated, the hand can be disposed and easily replaced by 3D printing a new one. Alternatively, since no electric or metal components are in the hand, it can be cleaned or sterilized for re-use. Future work may further explore other filament materials that are specifically resistant to acidity, radiation or high temperatures. Also, reinforcing the filament with metallic materials may be considered to improve durability but require optimal balance with proper hand flexibility. To speed up printing time, printing with a rigid material can be considered while incorporating flexibility through a unique joint geometry. Due to the robust and strong design, the hand could be further used for precise in-hand manipulations as done in previous work \cite{Sintov2019, icra2020a, Belief2019,SintovL4DC2020}. In such a case, each finger would be independently connected to its own actuator. Future work could also consider an algorithm for automated design of a hand given a specific task or set of objects.

\bibliographystyle{IEEEtran}
\bibliography{ref}

\end{document}